\RequirePackage{fix-cm}
\documentclass[twocolumn]{svjour3} 
\smartqed  
\usepackage{graphicx}
\usepackage{graphicx}
\usepackage{amssymb}
\usepackage{latexsym}
\usepackage{graphicx}
\usepackage{epstopdf}
\usepackage{caption}
\usepackage{url}
\usepackage{xcolor}
\usepackage{amsmath}
\usepackage{color, colortbl}
\usepackage{romannum}
\usepackage{makecell}
\usepackage{lscape}
\begin{document}

\title{Deep Neural Networks for COVID-19 Detection and Diagnosis using Images and Acoustic-based Techniques: A Recent Review}



\author{Walid Hariri         \and
        Ali Narin 
}


\institute{Corresponding author $^*$ Walid Hariri \at
             Labged Laboratory, Badji Mokhtar Annaba University \\
              \email{hariri@labged.net}           
          \and
          Ali Narin \at
              Department of Electrical and Electronics Engineering, Zonguldak Bulent Ecevit University\\ 
							\email{alinarin@beun.edu.tr}
				}

\date{Received: date / Accepted: date}

\maketitle

\begin{abstract}
The new coronavirus disease (COVID-19) has been declared a pandemic since March 2020 by the World Health Organization. It consists of an emerging viral infection with respiratory tropism that could develop atypical pneumonia. Experts emphasize the importance of early detection of those who have the COVID-19 virus. In this way, patients will be isolated from other people and the spread of the virus can be prevented. For this reason, it has become an area of interest to develop early diagnosis and detection methods to ensure a rapid treatment process and prevent the virus from spreading. Since the standard testing system is time-consuming and not available for everyone, alternative early-screening techniques have become an urgent need. In this study, the approaches used in the detection of COVID-19 based on deep learning (DL) algorithms, which have been popular in recent years, have been comprehensively discussed. The advantages and disadvantages of different approaches used in literature are examined in detail. The Computed Tomography of the chest and X-ray images give a rich representation of the patient's lung that is less time-consuming and allows an efficient viral pneumonia detection using the DL algorithms. The first step is the pre-processing of these images to remove noise. Next, deep features are extracted using multiple types of deep models (pre-trained models, generative models, generic neural networks, etc.). Finally, the classification is performed using the obtained features to decide whether the patient is infected by coronavirus or it is another lung disease. In this study, we also give a brief review of the latest applications of cough analysis to early screen the COVID-19, and human mobility estimation to limit its spread.
\end{abstract}

\section{Introduction}
\label{sec:intro}

\begin{figure}[h]
\centering
\captionsetup{justification=centering}
\includegraphics[width=.50\textwidth]{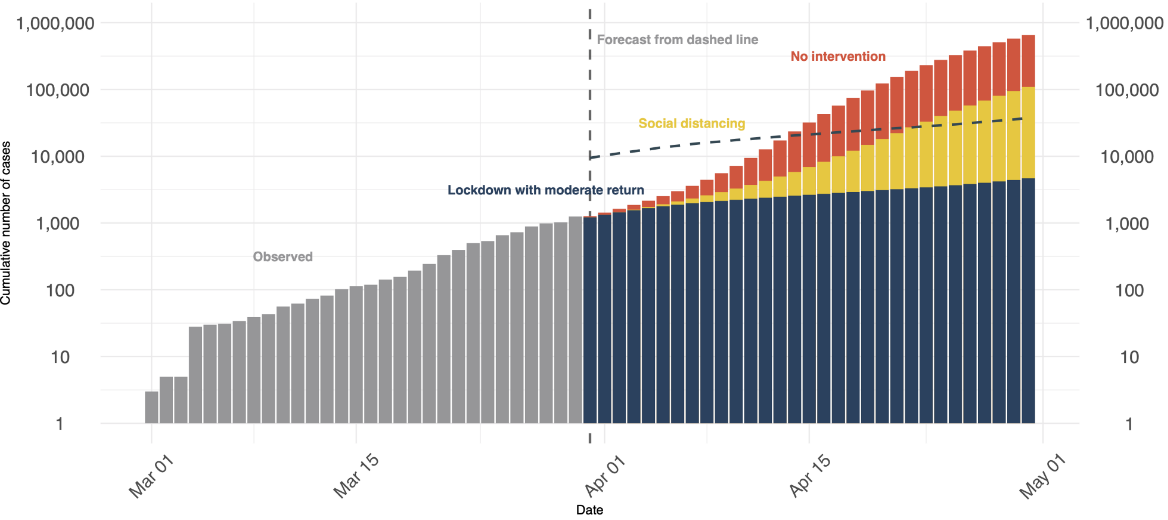}
\caption{Cumulative daily cases in India \cite{StudyGroup2021}.}
\label{fig:india_cases}
\end{figure}

The novel severe acute respiratory syndrome-related coronavirus (SARS-CoV-2) started from Wuhan, China in December 2019 and spread to all the countries worldwide. This virus caused pneumonia of unknown cytology and is named COVID-19. This infectious disease has been classified as a public health crisis of the international community concern on January 30, 2020, because of its high infectivity and mortality. In 2021, many variants of COVID-19 have been detected in different countries including UK, Brazil and India. The spread of these variants and and the mortality rates are more important than the previous ones. Figure \ref{fig:india_cases} presents the daily cases of the Indian variant reported by the COVID-19 study group in India. The lack of successful diagnosis or preventive measures has led to a rise in the number of cases, an increase in the cost of hospitalizations and palliative treatments. Therefore, scientists and medical industries around the world incited to find a prompt and accurate detection of COVID-19 for early prevention, screening, forecasting, drug development, and contact tracing to save more time for the scientific community and healthcare expert to pass to the next diagnosis stage to reduce the death rate reverse transcription polymerase chain reaction (RT-PCR) is recommended to diagnose COVID-19. Additionally, there are studies in the literature using various imaging methods (computed tomography (CT) and X-ray). \cite{xu2020deep,gozes2020rapid,ucar2020covidiagnosis,sethy2020detection,zhang2020covid}. It may occur in situations that negatively affect these methods. The changes of viruses by the appearance of new mutations make the classifications a more challenging task \cite{grubaugh2020making}. Moreover, one of the biggest problems with COVID-19 patients is viral pneumonia (VP). Differentiating between viral and non-viral pneumonia (nVP) is not easy. Coexistence of COVID-19 and viral pneumonia can have dire consequences.

Oxford COVID-19 Evidence Service Team Center follows some tips in identifying these problems. Muscle pain, loss of sense of smell and shortness of breath without pleuritic pain are the most common symptoms, especially in the case of COVID-19 infection. On the other hand, symptoms such as bilateral positive lung findings, tachycardia or tachypnea disproportionate to temperature, and low temperature indicate VP (not COVID-19) symptoms \cite{Carl:2020}. nVP, however, is most susceptible if it becomes rapidly unwell after a short period from the appearance of symptoms and does not have similar symptoms of COVID-19, pleuritic pain, or purulent sputum. 

\begin{figure}[h]
\centering
\captionsetup{justification=centering}
\includegraphics[width=.47\textwidth]{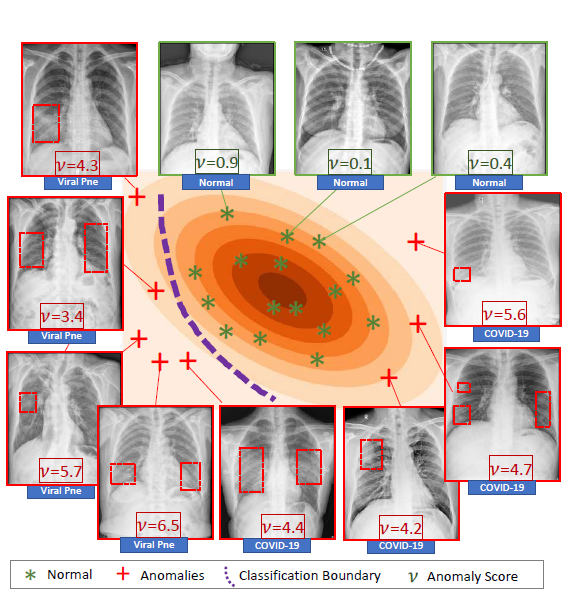}
\caption{Distinguishing between VP cases (anomalies) and nVP cases and normal controls \cite{zhang2020viral}.}
\label{fig:viral_pne}
\end{figure}

Many studies have been introduced to solve this problem, for instance, Zhang et al. \cite{zhang2020viral} proposed to lessen the process of anomaly detection into a one-class-classification problem using a confidence aware module. Deep learning (DL) is then used for the classification task as shown in Figure \ref{fig:viral_pne}. Recent reviews show that the use of novel technology with artificial intelligence (AI) and machine learning (ML) techniques considerably improves the screening, contact tracing, forecasting, and drug and vaccine development with high reliability.

Since the COVID-19 pandemic started, it has been clear that deep learning algorithms from ML technologies seem to be used extensively to detect COVID-19, VP, bacterial pneumonia (BP) and other similar cases. The advantages and disadvantages of these studies should be evaluated. In this study, it is aimed to present a detailed review on studies using DL approaches using various images in the literature to detect COVID-19. In addition, studies that detection of COVID-19 using acoustic sound data are included.

The rest of this paper is organized as follows: Section 2 presents the medical imaging technologies. In Section 3 we review the most important DL methods proposed to diagnose the COVID-19, as well as the recent advanced applications. Section 4 presents some additional DL applications to fight against COVID-19 such as acoustic analysis and human mobility estimation. In Section 5 an overall discussion and proposed solutions have been presented to accurately diagnose and to reduce the spread of the COVID-19. Conclusions, future trends and challenges end the paper.

\section{Medical imaging technologies versus RT-PCR test}
The medical imaging field has considerably emerged in the last years offering reliable automated methods for clinical decision making. It has received wide acceptance by the scientists and the medical community. In the case of COVID-19, CT scans and X-ray images can play a vital role in the early diagnosis of the disease. Infected patients have clinical symptoms including cough and fever, however, an important proportion of infected patients can be asymptomatic. In Germany, it has been confirmed in the study of Rothe et al. \cite{rothe2020transmission} that an asymptomatic patient was able to transmit the virus to another patient. According to the study of Al-Tawfiq et al. \cite{al2020asymptomatic} from 9 countries, 18 from 144 cases were asymptomatic, the equivalent of 12.5\%. The study has been done using the RT-PCR test. 

Due to the high risk of transmission of COVID-19, accurate diagnostic methods are urgently needed to prevent the spread of the virus and for humanity to breathe comfortably. Besides being the gold standard of the RT-PCR test, the results are time consuming (requires 5 to 6 hours) to obtain.In addition, the high rate of false detection of RT-PCR test is questioned whether it is a good diagnostic method.  \cite{xie2020chest,long2020diagnosis}. In this case, it is recommended that patients with typical imaging findings should be separated ones from one another and more than one RT-PCR test should performed to avoid misdiagnosis.

\begin{figure}[h]
\centering
\captionsetup{justification=centering}
\includegraphics[width=.48\textwidth]{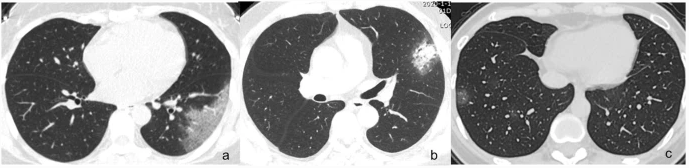}
\caption{ CT scans in the early fast gradually stage of COVID-19 pneumonia cases. a: GGO plus reticular pattern on the forth day. b: GGO plus consolidation on the third day. c: GGO on the second day. \cite{zhou2020imaging}}
\label{fig:CT_scan}
\end{figure}

The X-ray, however, is an efficient screening method, it is fast at capturing, cheaper than the RT-PCR test, and largely available worldwide. CT-scans, on the other hand, can be obtained much faster and more accurately in the presence of an efficient algorithm (notably DL algorithms) to accurately identify the infected patients. In \cite{liu2019comparison}, it has been proven that DL offers highly promising results for medical diagnostics compared to health-care professionals. Figure \ref{fig:CT_scan} presents the change that occurs in the COVID-19 pneumonia cases on some days. In the following section, detailed information about CT and X-ray images is presented.

\subsection{Chest Computed Tomography}
\label{sec:ct}
CT is an imaging method that uses a special x-ray beam to create detailed scans of areas inside the body (e.g. lungs, heart, blood vessels, airways, and lymph nodes). These images are taken from different angles to generate tomographic images which give the possibility to the radiographers to directly see inside the body instead of surgery. CT  images are considered to be an effective way of making clinical decisions. They showed high efficiency in diagnosing COVID-19 especially patients with false-negative RT-PCR results, assuming a role for the CT as a reliable tool for COVID-19 diagnosis during this epidemic period \cite{li2020coronavirus,ai2020correlation,xie2020chest,huang2020use}. Therefore, the National Health Commission of the People's Republic of China suggested CT examination in monitoring disease progression and controlling treatments of COVID-19 in its $6^{th}$ version of the diagnosis and treatment program \cite{zhao2020interpretations}. 
\subsection{X-Ray Image} 
\label{sec:xray}
\begin{figure}[h]
\centering
\captionsetup{justification=centering}
\includegraphics[width=.34\textwidth]{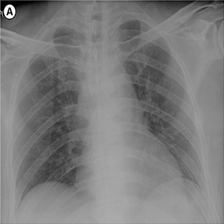}
\caption{An example of X-ray image for a COVID-19 patient.}
\label{fig:COVID_scan}
\end{figure}

Wilhelm Conrad Röntgen has discovered the first X-ray in 1895 during experimenting with Lenard tubes and Crookes tubes. X-ray has a very important role in the medical field, it can help in the prevention of infection, diagnosis, and control. X-ray scans are used worldwide to diagnose the injured part and for the detection or other diseases in order to treat patients \cite{ghosh2018automatic}. The X-ray facility is available even in the remotest parts and thus X-ray images can be easily acquired for patients even in their home or in their quarantine location. These images have been extensively used for COVID-19 diagnosis \cite{narin2020automatic}. The most common reported abnormal in Chest X-ray (CXR) findings are ground-glass opacities (GGOs) \cite{yoon2020chest}. Figure \ref{fig:COVID_scan} presents an example of an X-ray scan for COVID-19 patients.
CXR is the most widely used imaging technology by researchers because it is easily available and inexpensive. However, GGOs are often the first sign of a diagnosis of COVID-19 pneumonia.

\begin{figure*}[h]
\centering
\captionsetup{justification=centering}
\includegraphics[width=.80\textwidth]{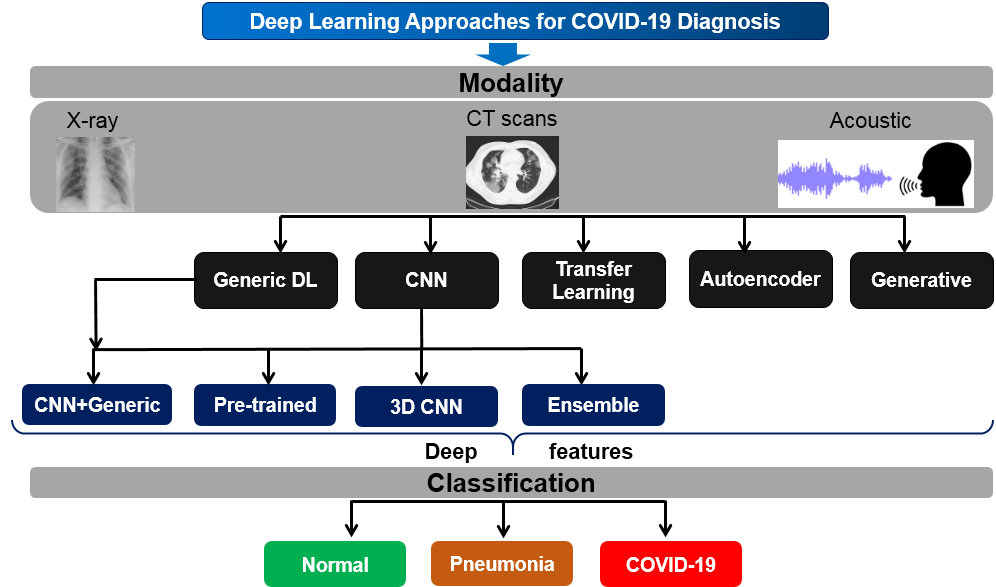}
\caption{Taxonomy of deep learning based-approaches for COVID-19 diagnosis.}
\label{fig:tax}
\end{figure*}

\section{Deep learning approaches in the COVID-19 pandemic} 
\label{sec:depp_m}
DL is a subset of  ML that offers considerable power for improving the accuracy and speed of diagnosis by automating the screening through medical imaging in collaboration with radiologists and/or physicians. Subsequently, it has received wide acceptance and interest by the medical community leads to emphasizing the development of such diagnostic technologies \cite{liu2019comparison}. In the following, we review the most important DL approaches adopted to diagnose the pneumonia of COVID-19 since its spread in December 2019 until today. Figure \ref{fig:tax} presents the taxonomy of these approaches using different images and acoustic features. In the following, we detail each of the distinguished nine groups:

\subsection{Generic deep learning} 
Generic DL methods without any specific modification have been proposed to detect COVID-19. For example, Wang el al. \cite{wang2020fully} have used CT images of 5,372 patients from 7 different cities in China to train a deep neural network (DNN). 

Pneumonia Detection Challenge dataset (RSNA) is used in \cite{luz2020towards} to train a DL model in order to locate lung opacities on chest radiographs. RSNA dataset contains two classes: Normal and Pneumonia (non-normal). The total of 16,680 images have been used from this data set where 8,066 are from healthy class (normal), whereas 8,614 as classified as pneumonia.

The authors in \cite{song2020deep} collected from two hospitals of in China the CT images of 88 infected patients (COVID-19), 101 patients diagnosed with bacteria pneumonia, where the rest are healthy (86 persons). Using this dataset, they applied a DL-based CT diagnosis system namely: \textit{DeepPneumonia} to localize the principal lesion features, especially GGO and thus to identify the infected patients. The first step is the segmentation of the lung region. Next, they introduced the DRE-Net (Details Relation Extraction neural network) to draw the top-K features in the CT images and to receive the image-level predictions. Finally, the image-level predictions is used to diagnose the patient.

Another generic DL framework is proposed in \cite{zhang2020automated} to automatically extract and analyze regions with high possibility to be infected with COVID-19. To do so, the authors applied a segmentation stage using a DL-based technique. Then, the infected regions were processed and quantized using specific metrics in the CT scan.

We can also find generic convolutional neural networks in which the authors use the generic CNN trained with their datasets without any combination with other ML algorithms or pre-trained models. For example, in \cite{fu2020deep}, the authors trained the CNN model with the data collected from Wuhan Jin Yin-Tan hospital in order to classify the CT images into one of the five following classes:  healthy lung, COVID-19, pneumonia, non-COVID-19 VP, BP and pulmonary tuberculosis.

\subsection{Transfer learning}
\begin{figure}[h]
\centering
\captionsetup{justification=centering}
\includegraphics[width=.48\textwidth]{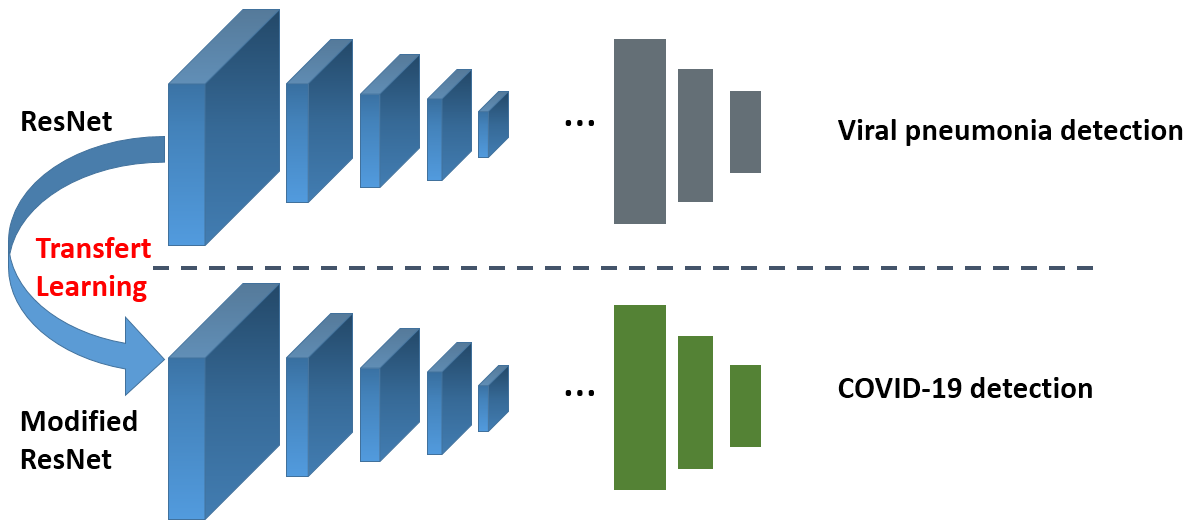}
\caption{An example of transfer learning process for COVID-19 detection.}
\label{fig:TL}
\end{figure}

Transfer learning (TL) is a ML technique in which a trained model for one task is redesigned in a related second task (see Figure \ref{fig:TL}). This approach is explicitly useful when there are not sufficient datasets like in the case of COVID-19 in order to either reduce the necessary fine-tuning data size or improve performance. TL can be used in two scenarios: supervised (with labeled data from the target domain) or unsupervised (without any labeled data from the target domain: the pretraining process is supervised, but unsupervised during fine-tuning).
A DNN is proposed in \cite{jaiswal2020covidpen} to detect COVID-19 using X-ray images. To do so, the authors applied a TL approach on the deep Pruned EfficientNet model. Then, it has been interpolated by post-hoc analysis to be able to explain the obtained predictions. TL based-framework for the detection of pneumonia is proposed in \cite{chouhan2020novel}. The features have been extracted from X-ray images using five different pre-trained models: DenseNet121, ResNet18, GoogLeNet, AlexNet and InceptionV3. Next,  an ensemble model has been added to combine outputs from all pre-trained models. The obtained results are as follows: accuracy of 96.4\%; recall of 99.62\% on non-trained data from the Guangzhou Women and Children's Medical Center database.

Fine-tuned deep TL with generative adversarial network (GAN) is presented in \cite{khalifa2020detection} to learn a limited dataset and to avoid the overfitting problem. To do so, the authors applied the pre-trained models: Squeeznet, AlexNet, GoogLeNet, and Resnet18 as deep TL models to detect pneumonia from chest x-rays. Applying a combination of GAN and deep transfer models enhanced the accuracy of the proposed system and realized 99\%. After applying image preprocessing algorithms to the chest X-ray images to identify and remove diaphragm regions, the pre-trained VGG-16 model \cite{simonyan2014very} has been fine-tuned in \cite{heidari2020improving} using the obtained images to predict COVID-19 infected pneumonia. Another work proposed in \cite{apostolopoulos2020covid} to detect the COVID-19 in small medical image datasets. To do so, they worked with two different data sets from public databases. In the first dataset, there are 224 COVID-19, 700 BP and 504 Normal X-ray images. The second dataset includes 224 COVID-19, 714 BP and VP, and 504 Normal X-ray images. They obtained 96.78\% accuracy, 98.66\% sensitivity and 96.46\% specificity performance values.

Multi-Channel TL-based method with X-ray images have been proposed in \cite{misra2020multi}. Multi-channel pre-trained ResNet model is then used to perform the diagnosis of COVID-19. To classify the X-ray images on a one-against-all strategy, three ResNet models have been retrained. The three allowed classifications are: 1) normal or diseased, 2) pneumonia or non-pneumonia, and 3) COVID-19 or non-COVID19 individuals. The method achieved a precision of 94\% and a recall of 100\%. Other TL-based methods can be found in \cite{minaee2020deep,maghdid2020diagnosing,benbrahim2020deep,haghanifar2020covid,abbas2020classification,rahaman2020identification,perumal2020detection,loey2021hybrid,pham2021classification}.

\subsection{Augmentation and Generation Techniques}
\begin{figure}[h]
\centering
\captionsetup{justification=centering}
\includegraphics[width=.50\textwidth]{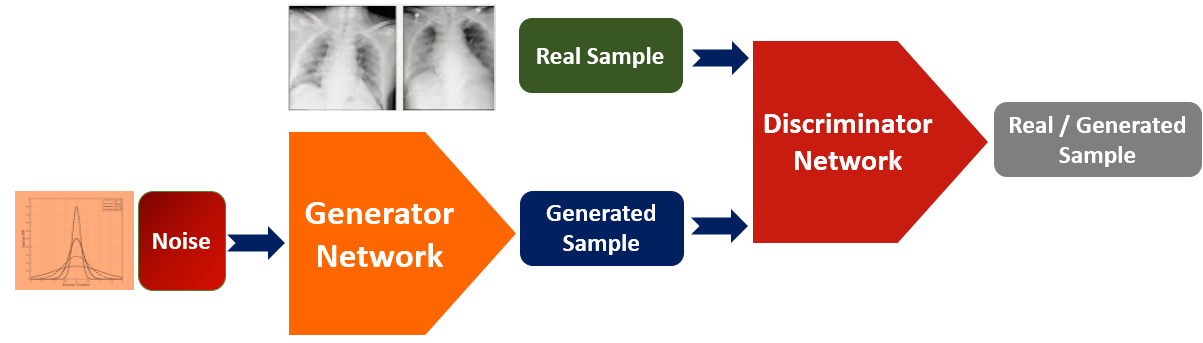}
\caption{Generative adversarial network representation for COVID-19 detection.}
\label{fig:GAN}
\end{figure}

Recently, generative adversarial networks (GANs) are considered the most powerful and successful method for data augmentation.
Since the outbreak of COVID-19 is recent, it is difficult to gather a significant amount of radiographic images and datasets in such a short time. Therefore, DL networks especially (CNNs) need additional training data to overcome this problem and to enhance the efficiency of CNN in detecting COVID-19 (See Figure \ref{fig:GAN}). Various methods have been applied the GANs for this reason. For instance, in \cite{waheed2020covidgan}, authors generate more X-ray images using Auxiliary Classifier Generative Adversarial Network (ACGAN) based on the CovidGAN model. Accordingly, the classification accuracy has been significantly enhanced from 85\% using the CNN alone, to 95\% using the ACGAN with CovidGAN model.

Also, to handle the problem of the lack of datasets for COVID-19, Loey et al. \cite{loey2020deep} proposed classical data augmentation techniques along with Conditional GAN (CGAN) on the basis of a deep transfer learning model for COVID-19 detection using CT images. Similar representation has been used in \cite{loey2020within} to classify the CT images into the following four classes : the COVID-19, normal, pneumonia bacterial, and pneumonia virus. To do so, the authors have used a dataset of 307 images. Three deep transfer models are then carried out in this work for investigation. The models are the Googlenet, Alexnet and Restnet18. three strategies have been conducted, in each strategy the authors applied a different deep TL using the three pre-trained models mentioned above. The testing accuracies achieved by the Googlenet, Alexnet and Restnet18 are 80.6\%, 85.2\% and 100\%, respectively. 

Another method proposed in \cite{karakanis2020lightweight} aims to generate synthetic medical images using DL CGAN to overcome the dataset limitation that leads to over-fitting. The proposed model has been implemented in a form to support a lightweight architecture without transfer learning without performance degradation. It can deal with any non-uniformity in the data distribution and the limited accessibility of training images in the classes. It consists of a single convolutional layer with filter size 32 and kernel 4 $\times$ 4, followed by ReLU activation function and Max Pooling layer for down-sampling the image (input representation) and enabling feature extraction. After a flatten layer there exists a dense layer of size 128, followed by dropout and a final dense layer with softmax activation function for a binary output. Other methods based on data augmentation to detect the COVID-19 can be found in \cite{ahmed2021convid}.

\subsection{Autoencoder-based models}
\begin{figure}[h]
\centering
\captionsetup{justification=centering}
\includegraphics[width=.50\textwidth]{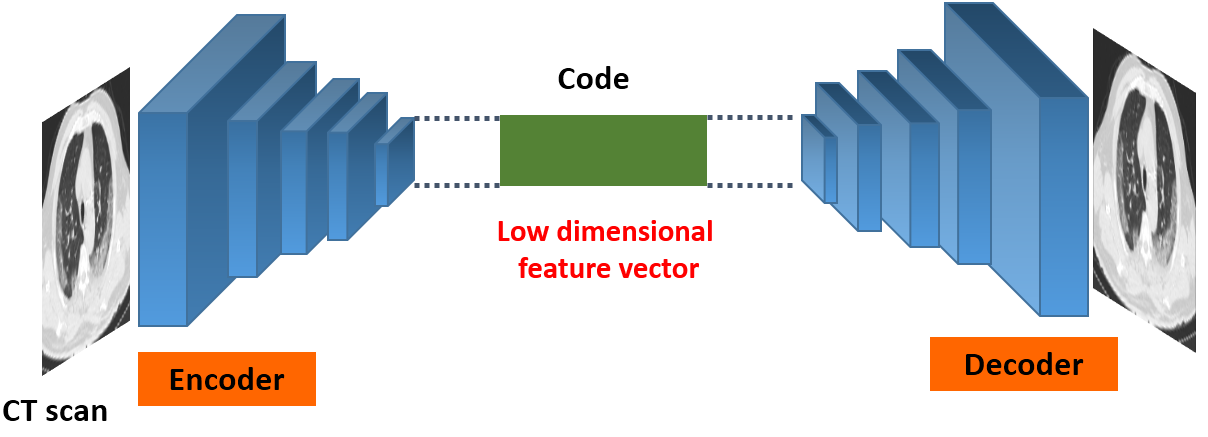}
\caption{An example of Autoencoder model for COVID-19 detection.}
\label{fig:encoder}
\end{figure}

Another ML technique to handle the problem of insufficient data for the affected COVID-19 cases is the Autoencoder (AE) \cite{baldi2012autoencoders}. It is a neural network method with competent data encoding and decoding strategies used for unsupervised feature learning. 
The AE models are comprised of two main steps: \textit{encoder} and \textit{decoder}. The input samples are mapped typically to a lower-dimensional space with beneficial feature representation in the encoding step. In the second step, the decoding consists of reverting data to its original space, trying to create data from lower space representation. Figure \ref{fig:encoder} shows the conceptual diagram of AE with its two main steps. The advantage of adopting such unsupervised classification to handle the problem of COVID-19 detection compared to its counterpart (supervised classification) is to avoid the long time spent in assembling large amounts of data which could increase the risk of mortality and postpones medical care. For example, in \cite{khoshbakhtian2020covidomaly}, the authors introduced the COVIDomaly which aims to diagnose new COVID-19 cases using a convolutional autoencoder framework. They tested two strategies on the COVIDX dataset acquired from the chest radiographs by training the model on chest X-rays: the first strategy used only healthy adults, the second tested healthy and BP, and infected adults with COVID-19. Using 3-fold cross-validation, they obtained a pooled Receiver Characteristic Operator-Area Under the Curve (ROC-AUC) of 76.52\% and 69.02\% with the two strategies respectively. 

In \cite{goel2020efficient}, the authors extracted discriminative features from the autoencoder and Gray Level Co-occurence Matrix using CT images. The obtained features are then combined with random forest classifier for COVID-19 detection. They achieved the following results:  accuracy of 97.78\%, specificity of 98.77\% and recall of 96.78\%. Other autoencoder-based methods can be found in \cite{berenguer2020explainable,shoeibi2020automated,khobahi2020coronet}.

\subsection{Pre-trained Deep Neural Networks}

Pre-trained models were originally trained on existing large-scale labeled dataset (e.g. ImageNet) and later fine-tuned over the chest CT and X-ray images to accomplish the diagnosis process. The last layer in these models has been removed and a new fully connected (FC) layer is added with an output size of two that represents two separate classes (COVID-19 or normal). In the obtained models, only the final FC layer is trained, while other layers are initialized with pre-trained weights \cite{nayak2020application}. These models can be a very useful solution to the lack of large datasets for COVID-19. However, some challenges exist. One of the risen problems here is that the transfer across datasets from a domain to another can lead to deterioration of performance due to the gap existing between the domains. This is often the case with medical images taken from different centers. Moreover, there is an over-fitting problem with small amounts of COVID-19 datasets. Therefore, pre-trained models are generally used with some particular modifications in order to avoid the over-fitting problem.

\begin{figure}[h]
\centering
\captionsetup{justification=centering}
\includegraphics[width=.50\textwidth]{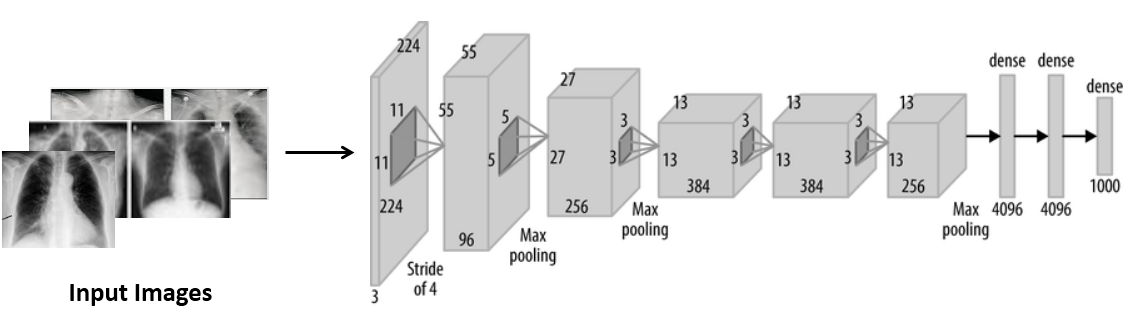}
\caption{AlexNet architecture proposed in \cite{krizhevsky2017imagenet}.}
\label{fig:alex}
\end{figure}

\begin{figure}[h]
\centering
\captionsetup{justification=centering}
\includegraphics[width=.50\textwidth]{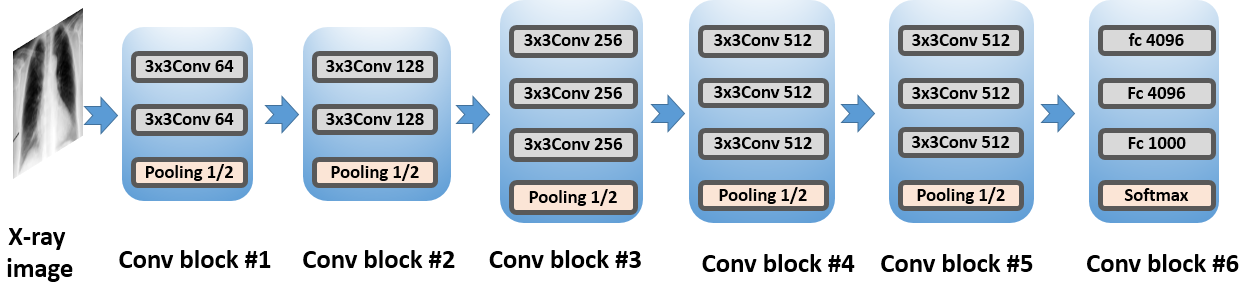}
\caption{VGG-16 architecture proposed in \cite{simonyan2014very}.}
\label{fig:vgg}
\end{figure}

\begin{figure}[h]
\centering
\captionsetup{justification=centering}
\includegraphics[width=.50\textwidth]{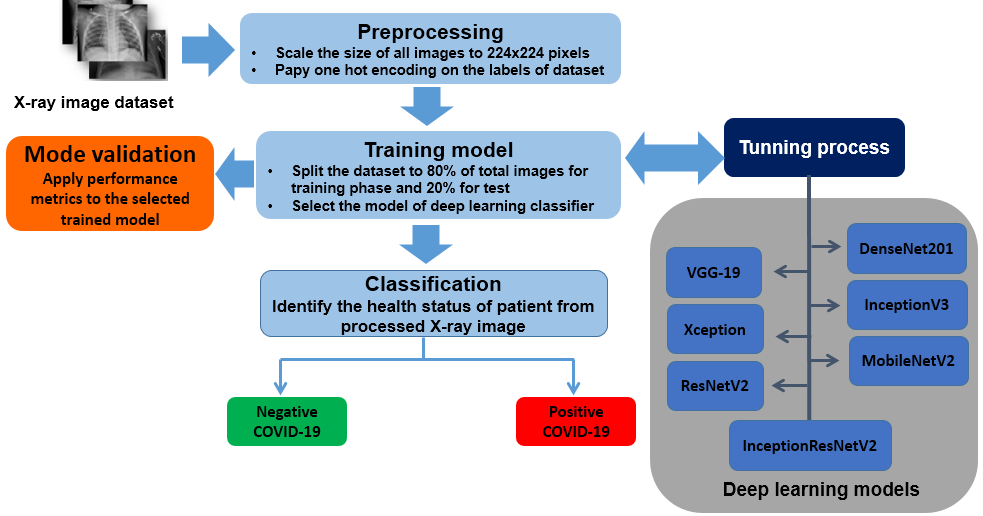}
\caption{COVIDX-Net framework \cite{hemdan2020covidx}.}
\label{fig:COVIDX}
\end{figure}

In \cite{nayak2020application}, eight pre-trained CNN models have been compared including GoogleNet, AlexNet, MobileNet-V2, VGG-16,   SqueezeNet, ResNet-50; ResNet-34 and Inception-V3. The obtained results for the classification of COVID-19 from normal cases show that  ResNet-34 outperformed the other pre-trained models and achieved an accuracy of 98.33\%. This evaluation has been conducted on a total of 286 scans of COVID-19 and normal classes as a training set, and 120 scans for the test (60 scans for each class). When dealing with the augmented dataset, the total of the training scans is 1002, where 428 scans are used for the validation and 120 scans for the test. ResNet-18 has been applied in \cite{oh2020deep} using limited training datasets and achieved a sensitivity of 100\% and 76.90\% of precision for the COVID-19 class.
Figures \ref{fig:alex} and \ref{fig:vgg} show an example of AlexNet and VGG-16 architectures respectively.
DenseNet201 pre-trained model is used in \cite{jaiswal2020classification} on chest CT images. To classify the patients into positive or negative COVID-19, deep transfer learning is carried out, and obtained a training accuracy of 99.82\%, and validation accuracy of 97.48\%.
The extreme version of the Inception (Xception) model is applied in \cite{das2020automated} and achieved an accuracy of 97.40\%, f measure of 96.96\%, sensitivity of	97.09\% and specificity of 97.29\% for three classes COVID-19, pneumonia, and other diseases. Hemdan et al. \cite{hemdan2020covidx} proposed COVIDX-Net framework to diagnose the COVID-19 cases using X-ray images. It includes three main steps to accomplish the diagnostic process of the COVID-19 as follows: pre-processing, Training Model and Validation, Classification. In consequence of the absence of public COVID-19 datasets, the experiments are carried out on 50 Chest X-ray images where only 25 have been diagnosed with COVID-19, for the validation. The COVIDX-Net combined the following seven different architectures of deep CNN models: VGG-19, DenseNet121, InceptionV3, ResNetV2, Inception-ResNet-V2, Xception, and the second version of Google MobileNet. Figure \ref{fig:COVIDX} presents an overview of the COVIDX-Net framework. Another model called  InstaCovNet-19 makes use of five pre-trained models including Xception, ResNet101, InceptionV3, MobileNet, and NASN is proposed in \cite{gupta2020instacovnet}. Two classification strategies have been conducted : (COVID-19, Pneumonia, Normal) and (COVID, NON-COVID). Very high precision and classification accuracy have been achieved using the two strategies (See Table \ref{tab:perf}). Similar method has been proposed by Narin et al. \cite{narin2020automatic} using five pre-trained CNNs and three three different binary datasets including COVID-19, normal (healthy), bacterial and viral pneumonia patients. Gour et al. \cite{gour2020stacked} proposed a new CNN model based on the VGG-19. They used a 30-layered CNN model for the training with X-ray images, and obtained sub-models using logistic regression. Other methods using pre-trained CNNs can be found in \cite{makris2020covid,afshar2020covid,kumar2020blockchain}.

\subsection{3D Convolutional Neural Networks}
3D CNN models have also been used in the literature. They mainly extract 3D features from the segmented 3D lung region using CT images. For example, Wang et al. \cite{wang2020weakly} segmented the lung region a pre-trained UNet model. The obtained volumes were fed into the proposed DeCoVNet (3D deep convolutional neural Network to Detect COVID-19). A weakly-supervised classification is then applied and achieved high COVID-19 classification performance and good lesion localization results. Muller et al. \cite{muller2020automated} have also used UNet model instead of computational complex CNNs to reduce the over-fitting problem during the segmentation of the infected lung region. In \cite{wang2020prior}, 3D-ResNet is applied for end-to-end training to classify the acquired lung images into pneumonia or healthy.

In order to predict the risk of COVID-19, Yang et al. \cite{yang2020end} applied end-to-end training from CT images using the 3D Inception V1 model pre-trained on the ImageNet dataset. The obtained accuracy was 95.78\% overall, and 99.4\% on a a part of the dataset containing 1,684 COVID-19 patients. Li et al. \cite{li2020artificial} introduced a 3D DL system that aims to early detect the COVID-19, called COVNet. The COVNet model is composed essentially of RestNet50, which have a range of CT scans as entry and produces features for the equivalent scans. The obtained features from all scans are then involved by a max-pooling process. The final feature map is used as an input to a fully connected layer and softmax activation function to produce an output of a likelihood result for the three classes: COVID-19, non-pneumonia and Community-acquired pneumonia (CAP). Han et al. \cite{han2020accurate} introduced a deep 3D multiple instance learning to detect the COVID-19 using CT images. High accuracy has been achieved (97.9\%) and AUC of 99.0\%. Other 3D CNN-based methods can be found in \cite{de2020improving,liu20203d}.

\subsection{Combination of Generic CNNs with traditional ML algorithms}
Another strategy is to use CNN models differently by combining them with traditional ML algorithms.

In \cite{stephen2019efficient}, the authors presented a CNN model trained on X-ray images to recognize pneumonia. The proposed architecture consists of a combination of the convolution, max-pooling, and rating layers. The obtained features comprise four convolutional layers, a max-pooling layer, and a RELU activator between them. The traditional ML algorithm ANN (Artificial neural network) is finally applied for classification. ANN and AlexNet architecture have been combined in \cite{aslan2020cnn} to systematically find out COVID-19 pneumonia subjects using CT scans. Firstly, a segmentation using ANN algorithm is performed to localize the lungs. Next, COVID-19 classes are augmented to produce more images. Finally, pre-trained Alexnet architecture is used in one time with only a transfer learning process, the obtained accuracy is 98.14\%. And with additional layer namely called (Bidirectional Long Short-Term Memories) in the second time, with an accuracy 98.70\%. Nour et al. \cite{nour2020novel} proposed a scratch CNN model including five convolution layered serial network. Three ML algorithms have been trained on the obtained deep features involving k-NN, SVM, and DT. The highest accuracy is obtained by the SVM with 98.97\%.

Instead of using pre-trained deep CNNs only as feature extractor, in \cite{ismael2020deep}, two other strategies have been conducted to accurately classify Chest X-ray images into positive of negative COVID-19 including \textit{fine-tuning strategy} and \textit{end to end training}. The following models have been used as a feature extractor : ResNet101, VGG19,  ResNet50, ResNet18, and VGG16 where SVM is used for ML-based classification. Whereas, a new CNN model is used  for the fine-tuning strategy. Finally, end-to-end training with a dataset of 180 COVID-19 and 200 normal is carried out as a third strategy. 94.7\% of accuracy is achieved using ResNet50 model and SVM classifier, where fine-tuned strategy with ResNet50 model achieved 92.6\%. Finally, the end-to-end training strategy of the developed CNN model realized a 91.6\% result. Deep CNN and long short-term memory (LSTM) have been combined in \cite{islam2020combined} to diagnose COVID-19 automatically from X-ray images. The obtained accuracy of the classification of three classes (COVID-19, normal, and other pneumonia) is 99.4\%. Similar methods that combine deep features and classical ML techniques can be found in \cite{sethy2020detection}. 

\subsection{Ensemble models}

Handling the problem of COVID-19 detection using a single DL model without any specific addition might not achieve a high accuracy classification using CXR images or CT scans. For this reason, the use of many DL models combined with each other can be a good solution, namely: ensemble model and the learning approach is called ensemble learning. For example, the authors in \cite{sitaula2020attention2} proposed a DL model namely Attention-based VGG-16. This model used VGG-16 to capture the spatial relationship between the ROIs in CXR images. By using an appropriate convolution layer (4th pooling layer) of the VGG-16 model in addition to the attention module, they added a novel DL model to perform fine-tuning in the classification process. In \cite{hall2020finding} ensemble of three pre-trained models including Resnet50 and VGG16 and an own small CNN is applied for a test set of 33 new COVID-19 and 218 pneumonia cases. The overall accuracy realized is of 91.24\%. Shalaf et al. \cite{shalbaf2020automated} an ensemble deep transfer learning system with 15 pre-trained CNN architectures on CT images. They obtained the following results: accuracy (85\%), precision (85.7\%) and recall (85.2\%).

\subsection{Smart phone applications}
To further automate the screening of COVID-19 and to make it faster, mobile phones can be a very interesting framework for that due to their facility and numerous sensors with important computing proficiencies. Specifically, a smartphone has is able to scan CT images of COVID-19 patients to use them for analysis screening. Moreover, multiple CT images of the same COVID-19 patient can be gathered into one smartphone for similarity examination of how disfigurement have been developed \cite{purswani2019big}. However, the computing capability of a mobile to treat a large amount of data is lower than a grand machine or a computer. Therefore, a lightweight representation is needed to accomplish this task. Consequently, various recent methods have been proposed to detect COVID-19 in mobile devices using a slight representation. In \cite{zulkifley2020covid}, a lightweight DL model namely \textit{LightCovidNet} has been offered to detect COVID-19 using a mobile platform. To enhance the performance of the proposed model, supplementary data have been generated and added to the training dataset using the conditional deep convolutional GAN. In order to reduce the memory usage of the proposed model, five units of feed-forward CNN are built using separable convolution operators. Multi-scale features are then learned to be suitable for the X-ray images which have been acquired from all over the world separately. Instead of COVID-19 diagnosis and detection, various lightweight applications have been introduced to delay the spread of the virus. These applications could be designed to be compatible with the capabilities of a smartphone to further speed up their operation. Among these applications we can find: masked face recognition \cite{hariri2020efficient}, facial mask detection \cite{chen2020face,chua2020face}, social distance monitoring \cite{ahmed2020deep,rezaei2020deepsocial} and human mobility estimation \cite{xiong2020mobile}. Other mobile-based technique using to fight against COVID-19 has been proposed in \cite{maghdid2020novel}. Using DL algorithms, the authors arrived to efficiently evaluate the level of pneumonia and thus to determine whether it is a COVID-19 case or not. 

\section{Deep learning for other applications}

Other applications have been proposed to fight against COVID-19 induced pneumonia, the most important are cough detection and human mobility estimation.
\subsection{Cough Detection}

In addition to the DL approaches using X-ray and chest CT scans for COVID-19 detection, scientists affirm that audio sounds generated by the respiratory system can be diagnosed and analyzed to decide whether the patient is infected or not. Therefore, cough analysis has been used to screen and diagnose COVID-19. ML techniques can supply useful cues enabling the development of a diagnostic instrument. To do so, cough data of COVID and non-COVID is required. Accordingly, Sharma et al. \cite{sharma2020coswara} proposed a database called Coswara, of respiratory sounds, namely, cough, breath, and voice. Some experiments have been recently carried out to screen COVID-19 from acoustic features, for example, in \cite{hassan2020covid}, Recurrent Neural Network (RNN) has been used in its new architecture, namely the Long Short-Term Memory (LSTM) to extract six speech features from a collected dataset (i.e. Spectral centroid, Spectral Roll-off, Zero-Crossing Rate, Zero-Crossing Rate, MFCC, and MFCC). In this work, 70\% of the data was used for training, and 30\% for testing. The obtained results show that the best accuracy
is achieved for breathing sound, reaching up to 98.2\% followed by cough sounds, an accuracy of 97\% is attained. Whereas, the accuracy of voice analysis is of 88.2\%.

\begin{figure}
\centering
\captionsetup{justification=centering}
\includegraphics[width=.50\textwidth]{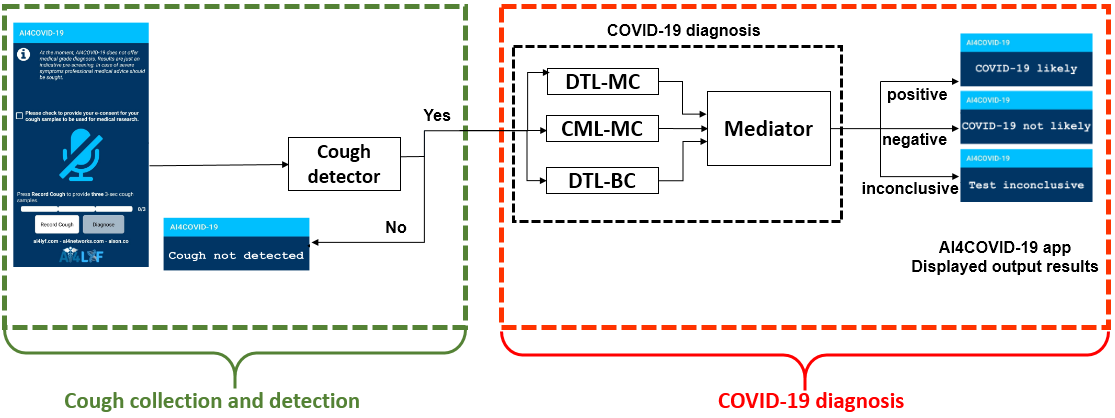}
\caption{AI4COVID-19 system architecture \cite{imran2020ai4covid}.}
\label{fig:AI4}
\end{figure}

A smartphone application using cough-based diagnosis for COVID-19 detection is proposed in \cite{imran2020ai4covid}. This application is based on an AI-powered screening solution called AI4COVID-19. Its principle is to send three 3 second cough sounds to an AI engine running in the cloud, and give a result during two minutes. To overcome the lack of COVID-19 cough training data, the authors applied transfer learning using ESC-50 dataset \cite{piczak2015esc} that contains 50 classes of cough and non-cough sounds acquired using a smartphone. Figure\ref{fig:AI4} presents the offered system architecture and a drawing of AI4COVID-19. The obtained results show high overall accuracy of 95.60\%, a sensitivity of 96.01\%, a specificity of 95.19\%, and precision of 95.22\%. Schuller et al. \cite{schuller2020covid} studied what computer audition could possibly contribute to the ongoing battle against the COVID-19. Other recent acoustic analysis for the detection of COVID-19 can be found in \cite{deshpande2020audio,alsabek2020studying,pal2020pay,quatieri2020framework,laguarta2020covid,pal2020pay,deshmukh2020interpreting}.

\subsection{Estimating Human Mobility}
Human mobility (movement) is one of the main factors that promote the transmission of the virus. Policy-makers find huge difficulties to find an optimal protocol to insure the social distancing and barrier measures. To solve this problem, Bao et al. \cite{bao2020covid} proposed a system that aims to evaluate and estimate maps of people movement responses by learning from existing ground truth data. The proposed system is based on a DL based-data generation called COVID-GAN. It merges a diversity of features involving contextual features, COVID-19 details and data history, as well as policies from various origins such as news, reports and SafeGraph. Experiment results showed that COVID-GAN can well imitate real-world human movement reactions and the area-constraint-based correction can considerably upgrade the solution value. To further explain the relation between people mobility and COVID-19 contamination, Xiong et al. \cite{xiong2020mobile} presented a study using mobile device data to give more insights to decision makers about the national mobility tendencies before and during the pandemic. 


\section{Discussion}
Table \ref{tab:perf} includes some studies conducted with DL models. It can be seen in Table 1 that studies have focused on two popular images, namely CT and X-ray images. The most used of these is X-ray images. This is because it is easily available everywhere. In addition, both its low memory space and high results in its performance encouraged researchers to use X-ray images. In addition, the greater availability of X-ray data from COVID-19 patients in public databases has led to the large number of these studies. In articles in the field of medicine, it is often stated that CT images show higher performance. However, these high accuracies are not seen in DL based CAD systems. This may be because radiologists can easily distinguish patients from CT images. CT images have too many cross-sections of the same person. They are much more complex than X-ray images. This complex situation is a disadvantage in distinguishing them in DL methods. The combination of X-ray and CT images, performed in few studies, also shows that it gives good results. It can be said that studies are carried out with multi-class solutions rather than the binary classification problem. It is seen that studies carried out with 3D data have lower performance. It has mostly been studied with 2D data. The results obtained from these data are very high.
Considering the number of data, it is known that DL models work stably with a lot of data. According to this fact, the number of data used is not sufficient. It can be stated that this is one of the biggest problems in the detection of COVID-19. It is very important to compare the studies conducted with the data obtained from many different centers. Otherwise, the accuracy of the studies can be fooled. For this, we recommend that data collected from many different centers be offered to researchers from a single center. However, in studies conducted with a small number of data, it is seen that pre-trained models are used to ensure high model training. Especially Keras, Tensorflow, PyTorch and MATLAB also include these pre-trained models. The biggest problem here is that these models are trained with the ImageNet dataset. Having a lot of data belonging to very different classes in the ImageNet dataset can reduce the trust in these models. However, it is seen that the performances obtained are also very high.

Traditional ML algorithms are also used by using the feature extraction part of DL models. This approach appears to improve performance. It is generally seen that the SVM algorithm is used. Most of the studies do not use any cross-validation (CV) method. We think that this decreases the reliability of the results. Because it is not known how the test data are distinguished. A very high performance test dataset can be created. This overestimates the results obtained. It can be noted that a CV method should be used in which the whole data set can be tested in studies. Although there are many studies using DL-based methods, it is very difficult to produce sufficiently transparent, stable and reliable models. It was clearly stated above that there are many parameters affecting the results.

As an alternative to the studies performed on X-ray and CT images, researches on the detection of COVID-19 are carried out with sound and cough-based acoustic sound analysis. Alternative approaches are very important in the detection of COVID-19. It may be possible to save more lives with real-time detection and diagnosis systems with online scanning systems installed with mobile or computer. As a result of all these approaches, namely RT-PCR test, DL-based X-ray and CT images, detection and acoustic examinations, it is predicted that patients with COVID-19 can be detected more stable and with higher accuracy. Because, considering that the epidemic started with a person, it is very important to correctly diagnose even a person.

\begin{landscape}
\begin{table}[p]
\captionsetup{justification=centering}
\caption{Summary of DL-based methods for COVID-19 pneumonia classification. \\ C.V: refers to Cross validation. Tr: refers to training, Val: refers to the validation, AE refers to Autoencoder. Sens refers to the Sensitivity. Spec refers to the Specificity. Acc refers to the Accuracy.}

\def\arraystretch{1.8}
\centering
\scriptsize
\begin{tabular}{l l l l l l l l l l l l|}\hline      
\textbf{Author}&\textbf{Modality}&\textbf{Dataset}&\textbf{2D/3D}&\textbf{All data}&\textbf{All COVID-19}&\textbf{Network and technique}&\textbf{C.V}&\textbf{Sens(\%)}&\textbf{Spec(\%)}&\textbf{Acc (\%)}\\ \hline 
\cite{narin2020automatic}&CXR&\makecell[l]{COVID-19/normal \\ COVID-19/pneumonia\\COVID-19/bacterial}&2D & \makecell[l]{ 3141\\1834  \\ 3113} &341& 5 pre-trained CNNs &5 & / & / &\makecell[l]{96.1 \\99.5 \\99.7 }\\ \hline
\cite{wang2020fully}&CT&COVID-19/pneumonia/normal  & 3D &1,266  &924  &DNN&/&\makecell[l]{Tr: 78.93\\Val: 80.39}&\makecell[l]{Tr: 89.93\\Val: 81.16}	&/ \\  \hline      
\cite{song2020deep}&CT& COVID-19/normal/bacterial & 2D &275  &88 &DRE-Net&/&93&96&99\\ \hline   
\cite{khalifa2020detection}&CXR&pneumonia dataset  & 2D &624 &50  &GAN + TL&/&/&/&99\\ \hline   
\cite{apostolopoulos2020covid}&CXR&COVID-19/normal/bacterial &2D & 1,427 &224&MobileNet v2&10&98.66&96.46&94.72\\ \hline
\cite{misra2020multi}&CXR&COVID-19/pneumonia/normal &2D & 6,008 &184&Three ResNet models&5& / & / &93.9\\ \hline    
\cite{khoshbakhtian2020covidomaly} &CXR&COVID-19/pneumonia/normal &2D & 8,850 &498&AE : COVIDomaly&3& / & / &76.52\\ \hline
\cite{nour2020novel}&CXR&COVID-19/pneumonia/normal &2D & 2,905 &219& CNN+k-NN+SVM &/ & / & / &98.70\\ \hline   
\cite{aslan2020cnn}&CXR&COVID-19/pneumonia/normal &2D & 2,905 &219& ANN+AlexNet &/ & / & / &98.97\\ \hline  
\cite{oh2020deep}&CXR&COVID-19/pneumonia/normal &2D & 502 &180& ResNet-18 &/& 76.90 & 100  &/ \\ \hline 
\cite{hall2020finding}&CXR&COVID-19/pneumonia/normal &2D & 2,905 &219& \makecell[l]{Ensemble:Resnet50 \\and VGG16} &10 & / & / &91.24\\ \hline  
\cite{jaiswal2020classification}&CXR&COVID-19/normal &2D & 2,492 &1,262& TL and DenseNet201 &/ & / & / &99.82\\ \hline
\cite{das2020automated}&CXR&COVID-19/pneumonia/normal &2D & / & / & Xception &/ & 97.09 & 97.29 &97.40\\ \hline
\cite{ismael2020deep} &\makecell[l]{CXR}&COVID-19/normal &2D & 380 & 180 & \makecell[l]{5 pre-trained models\\+ SVM} &/ & / & / &94.7\\ \hline                        
\cite{gupta2020instacovnet}&\makecell[l]{CXR}& \makecell[l]{COVID-19/pneumonia/normal\\COVID/non-COVID} &2D& 2905 & 219  &  pre-trained models  &/ & / & / &\makecell[l]{99.08\\99.53  }\\ \hline   
\cite{makris2020covid}&CXR&COVID-19/pneumonia/normal &2D & / & / & 5 pre-trained CNNs &/ & / & / &95\\ \hline
\cite{afshar2020covid}&CXR&\makecell[l]{Bacterial, Non-COVID Viral, \\COVID-19} &2D & / & / & 5 COVID-CAPS  &/ &  90 & 95.8 &95.7\\ \hline
\cite{minaee2020deep}&CXR&COVID-19/normal &2D & 5,000 & 184 & 5 TL+pre-trained models &/ & 100 & 98.3 &/\\ \hline
\cite{maghdid2020diagnosing}&CXR+CT&COVID-19/normal &2D & 526 & 238 &  TL+AlexNet model &/ & 72 & 100 &94.1\\ \hline
\cite{benbrahim2020deep}&CXR+CT&COVID-19/normal &2D & 320 & 160 & \makecell[l]{TL+InceptionV3 \\ and ResNet50} &/ & 72 & 100 &99.01\\ \hline
\cite{islam2020combined}&CXR&COVID-19/pneumonia/normal&2D & 4,575  & 1,525  &  LSTM+CNN &/ & 99.2 & 99.9 &99.4\\ \hline
 \cite{yang2020end}&CXR&COVID-19/pneumonia/normal&2D & \makecell[l]{4,448\\101}  & \makecell[l]{2,479\\ 52} & 3D Inception V1 &10 &\makecell[l]{ / \\98.08}& \makecell[l]{/ \\91.30\\}&\makecell[l]{95.78\\ 93.3}\\ \hline
\cite{zulkifley2020covid}&CXR&COVID-19/pneumonia/normal &2D & 1343 & 446 & \makecell[l]{Conditional GAN : \\LightCovidNet} &5 & / & / &97.28\\ \hline
\cite{heidari2020improving}&CXR&COVID-19/normal &2D & \makecell[l]{Total: 8,504 \\Training: 6,899}	& \makecell[l]{Total: 445 \\Training: 366} & TL VGG-16 model&/ &98.0 & 100 & 94.5 \\ \hline
\cite{shalbaf2020automated}&CT&COVID-19/normal &2D & 746 & 349 & \makecell[l]{TL+ Ensemble of \\15 pre-trained models} &/ & / & / &85\\ \hline
\cite{goel2020efficient}&CT&COVID-19/normal &2D & 2,482 & 1,252 & AE+random forest &/ & / & 98.77 &97.87\\ 
\hline

\cite{hemdan2020covidx}& CXR & COVID-19/normal  &3D  &50 &25&COVIDX-Net	&/&100	&80	& /\\ \hline 

\end{tabular}

\label{tab:perf}
\end{table}
\end{landscape}

\section{Conclusion}
Although the RT-PCR test is considered the gold standard for COVID-19 diagnosis, it is time-consuming to make a decision because of high false-negative levels in the results. Therefore, medical imaging modalities such as chest X-ray and chest CT scans are the best alternative according to scientists. Chest X-ray radiography is of low cost and low radiation dose, it is available and easy to use in general or community hospitals. This review presents a detailed study of the existing solutions that are mainly based on DL techniques to early diagnose the COVID-19. This study gives more of an insight into the scientists’ and decision-makers’ thought processes - not only during the wave periods but also during that of the vaccination that could require real-time mass testing. The lack of data, however, is the mandatory problem to achieve efficient and real-time results. Many solutions have been presented and discussed in this review study to give more ideas to future trends and also for eventual future diseases that might suffer from the missing-data problem. We believe that with more public databases, better DL based-approaches can be developed to detect and diagnose the COVID19 accurately. Also, when policy-makers and citizens are making their best to submit to the difficult constraints of lockdown and social distancing, AI can be used to create more intelligent robots and autonomous machines to help health workforce and to reduce their workload by disinfection, working in hospitals, food distributing and helping the patients. The challenge of this solution is that people lack confidence in autonomous machines and prefer to be served by a human even if there is a risk of virus transmission. Moreover, entrusting chatbots to diagnose patients needs a large amount of medical data from experts. Also, the difference in languages from a country to another makes an already difficult task still more arduous. On the other hand, when dealing with voice analysis, there are still many challenges to be taken up. For example, until now, annotated data of patients' voices are not publicly available for research purposes of COVID-19 detection and diagnosis. Collecting these data is mostly made in unconstrained environments (i.e. in-the-wild) using smartphones or other voice recorders. These environments are generally noisy and contain reverberation, which leads to bad quality of data and makes the diagnosis and detection of COVID-19 more challenging. Finally, one of the most important future trends is to concentrate on further decreasing the false negative rate and, as far as practicable, reducing the false positive rate by the same token to accurately differentiate viral from BP.

\begin{acknowledgements}
The authors would like to thank the 'Agence Nationale de Valorisation des Résultats de la Recherche et du Développement Technologique (DGRSDT), Algérie'.
\end{acknowledgements}

%
 \section*{Conflict of interest}
 The authors declare that they have no conflict of interest.



\bibliographystyle{spmpsci}
\bibliography{refs}

\end{document}